\documentclass[]{fairmeta}
\usepackage{amssymb}
\usepackage{lmodern}

\DeclareUnicodeCharacter{2212}{-}

\title{Polarization-Based Eye Tracking 

with Personalized Siamese Architectures}

\author[1]{Beyza Kalkanli}
\author[2]{Tom Bu}
\author[1]{Mahsa Shakeri}
\author[1]{Alexander Fix}
\author[1]{Dave Stronks}
\author[1]{Dmitri Model}
\author[1]{Mantas \v{Z}urauskas}

\affiliation[1]{Meta, Reality Labs, Redmond, WA 98052, USA}
\affiliation[2]{Meta, Reality Labs, Burlingame, CA 94010, USA}

\abstract{Head-mounted devices integrated with eye tracking promise a solution for natural human-computer interaction. However, they typically require per-user calibration for optimal performance due to inter-person variability. A differential personalization approach using Siamese architectures learns relative gaze displacements and reconstructs absolute gaze from a small set of calibration frames. In this paper, we benchmark Siamese personalization on polarization-enabled eye tracking. For benchmarking, we use a 338-subject dataset captured with a polarization-sensitive camera and 850 nm illumination. We achieve performance comparable to linear calibration with 10-fold fewer samples. Using polarization inputs for Siamese personalization reduces gaze error by up to 12\% compared to near-infrared (NIR)-based inputs. Combining Siamese personalization with linear calibration yields further improvements of up to 13\% over a linearly calibrated baseline. These results establish Siamese personalization as a practical approach enabling accurate eye tracking.}

\date{\today}
\correspondence{Mantas \v{Z}urauskas at \email{mantas@meta.com}}

\begin{document}

\maketitle

\section{Introduction}

Eye tracking (ET) enables critical applications in virtual/augmented reality~\cite{10.1007/s10055-022-00738-z}, human-computer interaction~\cite{CHEN2023883}, and assistive technologies~\cite{9745906}. In practice, achieving accurate gaze estimation must contend with substantial inter-person variability in eye anatomy and physiology, occlusions from eyelids and eyelashes, headset fit changes, and session-to-session drift. To deliver reliable interaction, most systems rely on per-user calibration.

Siamese personalization is an approach that learns relative gaze displacements between pairs of eye images from the same individual and reconstructs absolute gaze at test time using a small anchor set. Specifically, following the differential-gaze formulation of~\cite{8920005}, a shared-weight (Siamese) network predicts the offset between a test image and each calibration image; aggregating these offsets with the known anchor labels yields the final absolute gaze. This formulation turns calibration into a few-shot problem, leveraging pairwise relationships rather than requiring dense per-user regression. Importantly, the approach is modality-agnostic and naturally extends to polarization-enabled eye tracking (PET), which reveals person-specific scleral and corneal cues not visible in intensity-only imaging~\cite{pet_paper}.

Our application scenario uses binocular eye images and imposes strict calibration-budget constraints. In contrast to conventional linear regression calibration~\cite{8920005}, which typically requires tens to hundreds of on-device samples to reach competitive accuracy, Siamese personalization operates effectively with an order of magnitude fewer images. We further show that Siamese personalization is complementary to linear calibration: using Siamese predictions as the input to a lightweight linear post-calibration yields additional gains, providing a practical path to high-accuracy ET with minimal user effort.

Our main contributions are as follows:
\begin{itemize}

    \item We demonstrate that, Siamese personalization can effectively be applied to Polarization-enabled Eye-Tracking (PET)\cite{pet_paper}. We show consistent improvements with Siamese personalization across polarization-sensitive inputs, validating the effectiveness of this approach compared to non-calibrated or linearly calibrated Baseline models. 
    
    \item In addition, we demonstrate that Siamese personalization can be seamlessly combined with linear calibration. This combination reduces test gaze errors on PET compared to applying linear calibration to the Baseline model by 13\% at P50 (50th percentile/median), 11\%, at P75 (75th percentile) and 8.6\% at P95 (95th percentile) of the error distribution.
    
    \item Notably, Siamese personalization for PET achieves competitive performance on polarization data with 10× fewer calibration images compared to linear calibration, significantly improving practicality for real-world applications.

\end{itemize}

The rest of the paper is organized as follows: Section~\ref{sec:Related} reviews related work, Section~\ref{sec:Method} describes our approach, Section~\ref{sec:Experiments} presents the experimental setup and results, and Section~\ref{sec:Conclusion} concludes with discussion and future directions.

\section{Related Work}
\label{sec:Related}

Initial efforts in eye tracking (ET) include model-based and feature-based approaches that relied on geometric eye models and hand-crafted features~\cite{1634506, Valenti2012CombiningHP}. While these approaches offer interpretability and computational efficiency, they have significant limitations: they require careful feature engineering\cite{jemr.7.1.4}, rely on restrictive assumptions about eye geometry\cite{Swirski2013}, need extensive calibration to handle individual anatomical variations\cite{10.3389/fpsyg.2024.1309047}, and exhibit lower robustness to challenging imaging conditions including varying camera distances\cite{HANSEN2005155}, eyeglasses, and environmental changes~\cite{jemr.14.4.2}. 

Following advances in deep learning, the field evolved toward data-driven appearance-based methods~\cite{10508472}. Modern methods for ET use machine learning, primarily trained end-to-end to regress gaze direction directly from images of the eye~\cite{7299081}. These deep learning-based methods address the limitations faced by model-based approaches by learning representations directly from data, demonstrating significant improvements in both accuracy and robustness~\cite{Fischer_2018_ECCV, cheng2020coarse}.

To improve performance, some appearance-based approaches incorporate additional contextual information beyond eye regions. These methods adopt various strategies: some process full-face images to jointly capture eye appearance and head orientation~\cite{7050250, 8015018, o2022self}, while others combine full-face images with separately extracted eye regions or encoded head location information~\cite{7780608, 9008783}. Although multi-modal methods offer improved accuracy and robustness, they require additional sensors or larger fields of view~\cite{8015018}. Depending on application constraints, appearance-based methods that rely solely on eye images may be the only viable solution, particularly in scenarios with limited field of view such as near-eye head-mounted eye trackers. To address these constraints, several works have developed appearance-based methods that learn direct mappings from eye image appearance to gaze direction using various machine learning techniques~\cite{7299081}. These methods can be categorized by their input strategy: single-eye methods process left and right eyes independently~\cite{Zhang2020ETHXGaze}, while binocular approaches leverage complementary information from both eyes~\cite{kim2019}.

Regardless of input modality, physiological differences between individuals, including variations in eye anatomy, corneal shape, and pupil dynamics, introduce significant inter-subject variability~\cite{8003267}. This necessitates personalization or calibration to achieve accurate eye tracking~\cite{10.1145/3290605.3300646}. In model-based eye tracking, calibration is inherently integrated into the geometric modeling process~\cite{4770110}. For appearance-based methods, various personalization approaches have been proposed. Traditional post-processing calibration methods, such as linear calibration, use a sequence of frames to derive person-specific regression parameters that refine the base model's gaze predictions~\cite{aria}. Recent deep learning-based personalization methods explore domain adaptation and fine-tuning~\cite{Wang_2022_CVPR, 9157235}. While direct fine-tuning on person-specific data can yield strong results, it risks overfitting when calibration data is limited or adaptation strategies are suboptimal~\cite{50093}. Some studies separate feature extraction and regression components, adapting only the feature encoder to user-specific data~\cite{10.1145/3173574.3174198}, while others model personal variations as low-dimensional latent parameters per eye~\cite{9022231}. While personalization is crucial for achieving high accuracy across diverse users, minimizing calibration data requirements remains a key challenge. Recent work~\cite{50093} employs few-shot learning to reduce calibration requirements through embedding-based calibration from full face images, demonstrating that using fewer calibration images may require additional input modalities. Despite these advances, achieving high accuracy using only eye images with minimal calibration overhead remains an open challenge. To address this, this work specifically investigates personalization methods that can use fewer calibration images by leveraging the information contained in binocular PET images.

Siamese networks, which learn similarity metrics by processing pairs of inputs through shared-weight architectures, have proven effective for tasks such as image recognition~\cite{koch2015siamese} and object tracking~\cite{10.1007/978-3-319-48881-3_56}. In the context of gaze estimation, \cite{8920005} demonstrated that Siamese architectures can effectively learn user-specific gaze patterns by comparing eye images from the same individual across different time points. This differential gaze estimation approach predicts relative gaze displacements between image pairs rather than absolute coordinates, enabling more efficient use of limited calibration data by exploiting pairwise relationships. However, the potential of Siamese-based personalization for polarization-enabled eye tracking remains unexplored. Polarization-based methods provide richer input features than conventional intensity images~\cite{pet_paper}, which is especially beneficial when the iris/pupil is not visible and mostly the white sclera is visible in the captured images. Our work addresses this gap by demonstrating that Siamese network-based personalization can leverage polarization information to achieve improved accuracy with reduced calibration requirements.

\section{Method}
\label{sec:Method}
\subsection{Problem Definition}
Given a pair of eye images captured from left and right cameras, our goal is to estimate the user's gaze direction in polar angle coordinates. The eye tracking problem here can be formulated as a supervised learning problem in which a model learns a mapping $f: \mathcal{I} \rightarrow \mathbb{R}^4$. The input $\mathcal{I} = (\mathcal{I}^{\text{left}}, \mathcal{I}^{\text{right}})$ represents the left and right eye images with $\mathcal{I}^{\text{left}}, \mathcal{I}^{\text{right}} \in \mathbb{R}^{D \times H \times W}$, and the output is the predicted gaze angles $\mathbf{\hat{g}} = [\hat{\theta}_{\text{left}}, \hat{\varphi}_{\text{left}}, \hat{\theta}_{\text{right}}, \hat{\varphi}_{\text{right}}]$, where $\theta$ represents yaw and $\varphi$ represents pitch. Here, $D$ depends on how the polarization data is processed (e.g., different polarization channels or derived features), and $H, W$ depend on the downsampling factor. For example, $D=4$ when using raw polarization angles ($0^\circ, 45^\circ, 90^\circ, 135^\circ$), or $D=3$ when using derived features such as intensity, degree of linear polarization, and angle of linear polarization. The corresponding ground truth polar coordinates come from the gaze targets displayed to the user during data collection and denoted as: $\mathbf{g_{gt}} = [\theta_{\text{left}}, \varphi_{\text{left}}, \theta_{\text{right}}, \varphi_{\text{right}}]$. 

\subsection{Polarization Data.} 
\begin{figure}[htbp]
    \centering
    \begin{subfigure}[b]{0.3\textwidth}
        \centering
        \includegraphics[width=\textwidth]{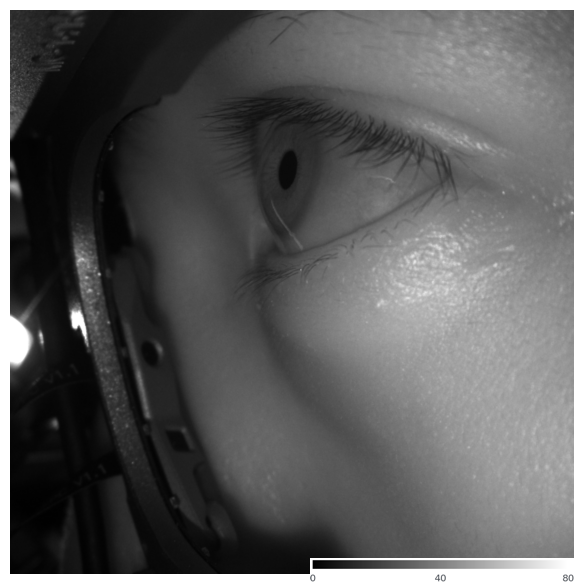}
        \caption{Intensity}
        \label{fig:intensity}
    \end{subfigure}
    \hfill
    \begin{subfigure}[b]{0.3\textwidth}
        \centering
        \includegraphics[width=\textwidth]{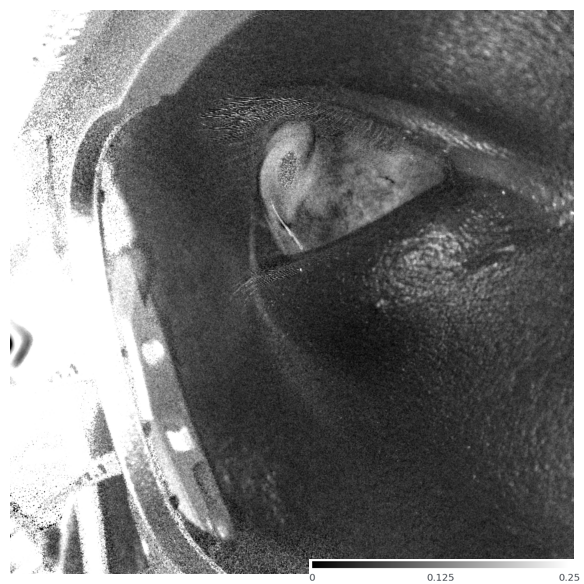}
        \caption{Degree of Linear Polarization (DoLP)}
        \label{fig:dolp}
    \end{subfigure}
    \hfill
    \begin{subfigure}[b]{0.3\textwidth}
        \centering
        \includegraphics[width=\textwidth]{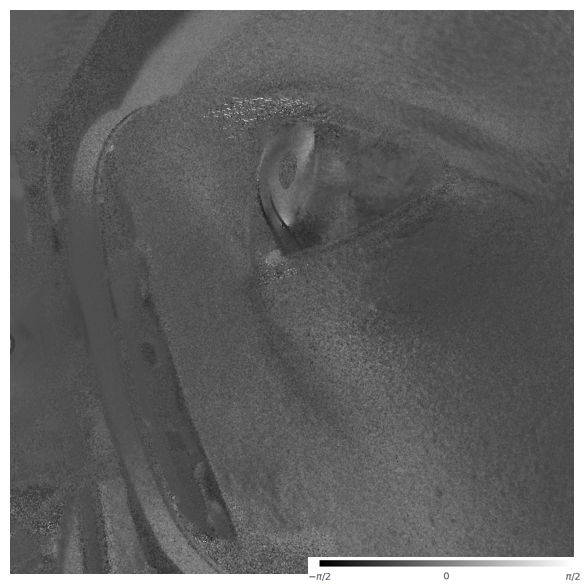}
        \caption{Angle of Linear Polarization (AoLP)}
        \label{fig:aolp}
    \end{subfigure}
    \caption{Intensity, DoLP, and AoLP channels derived from polarization-sensitive camera data.}
    \label{fig:data}
\end{figure}

Following the preprocessing approach in~\cite{pet_paper}, we compute Intensity, Degree of Linear Polarization (DoLP), and Angle of Linear Polarization (AoLP) from the polarization camera that captures raw intensities at four linear orientations ($0^\circ, 45^\circ, 90^\circ, 135^\circ$). After demosaicking and Gaussian smoothing ($\sigma = 1$), we compute Stokes parameters: $S_0 = I_{0^\circ} + I_{45^\circ} + I_{90^\circ} + I_{135^\circ}$, $S_1 = I_{0^\circ} - I_{90^\circ}$, and $S_2 = I_{45^\circ} - I_{135^\circ}$. From these, we derive Intensity $I = S_0/4$, $\text{DoLP} = \sqrt{S_1^2 + S_2^2}/(S_0 + \varepsilon)$ where $\varepsilon$ ensures numerical stability, and $\text{AoLP} = \frac{1}{2}\arctan2(S_2, S_1)$ as visualized in Figure~\ref{fig:data}.

Physically, DoLP quantifies the fraction of linearly polarized light,  while AoLP encodes the polarization orientation~\cite{pet_paper}. In  the sclera, anisotropic collagen under multiple scattering yields fine-scale, temporally stable texture largely absent in intensity imaging. At the cornea, Fresnel reflections at the tear film interface and birefringence from stromal lamellae produce characteristic polarization patterns. These subject-specific features persist over time~\cite{pet_paper}, making them well-suited for Siamese differential learning where the network leverages stable, person-specific anchors to estimate gaze displacements.

\subsection{Siamese Model for Personalization}
\label{sec:siamese}
\begin{figure*}[h]
  \centering
  \includegraphics[width=0.9\linewidth]{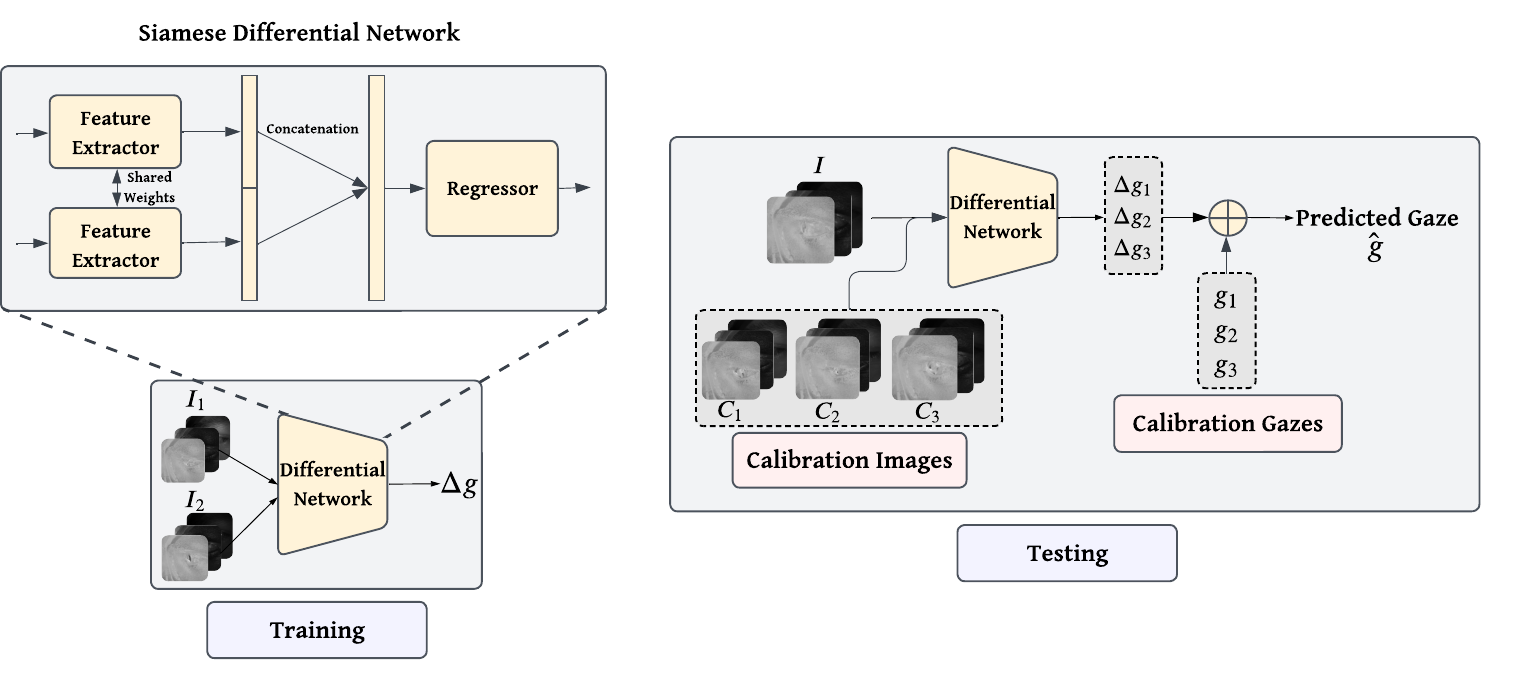}
  \caption{The polarization-based dataset is split into training and testing subjects. Both training and testing employ a Siamese network architecture. During training, the network learns to predict gaze displacement between pairs of images from the same eye of the same subject. During testing, the predicted gaze for an input image is computed by estimating gaze displacements between the input and each image in the subject's calibration set, then aggregating the calibration gaze labels based on the predicted displacements. While our approach processes binocular images as input to each Siamese branch, we illustrate with single-eye images for clarity.}

  \label{fig:siamese_pipeline}
\end{figure*}

The Siamese approach proposed in~\cite{8920005} tackles the eye tracking problem by learning the gaze displacement between images of the same eye from the same person. Rather than directly predicting absolute gaze coordinates, the Siamese network architecture learns a function $f: (I_1, I_2) \rightarrow \Delta g$ that maps two eye images to the relative displacement between their corresponding gaze targets. We adapt this approach to work with binocular image pairs. In our implementation, $I_1$ and $I_2$ denote binocular input pairs captured from the same person as $(I_{1}^{\text{left}}, I_{1}^{\text{right}})$ and $(I_{2}^{\text{left}}, I_{2}^{\text{right}})$, respectively, $g$ represents the gaze target, and $\Delta g$ is the gaze displacement between the two inputs.
The Siamese differential network consists of two branches with shared weights, each accepting a binocular eye image pair. Each branch extracts features from a left-right camera image pair, with each pair captured at a different time point from the same user. After concatenating the extracted features, the regressor predicts the gaze displacement $\Delta g$ between the two inputs. The backbone structure that is used for the branches of Siamese model is further explained in Section \ref{sec:Experiments}. Figure~\ref{fig:siamese_pipeline} provides a high-level overview of the training and testing processes. While both training and testing use the same differential network structure with gaze displacement as output, they differ in how this output is utilized. At test time, we additionally aggregate across multiple inputs, averaging the predictions of gaze displacements from all calibration images to obtain the final gaze.

\textbf{Training.} The model receives two inputs and predicts the gaze displacement $\Delta g$ between them. We use the \textit{random sampling} strategy from~\cite{8920005} for training data, which generates random pairs of inputs from the same subject, applied across all training subjects. The pairwise sampling of the training set generate more training samples from person-specific data. By exploiting the relative gaze displacement approach between images, the model can generalize effectively from sparse calibration examples, making it practical for real-world deployment. 

We use Smooth L1 loss with outlier rejection \cite{aria} as the loss function during training. For errors below threshold $\theta$, we apply standard Smooth L1 loss; for errors exceeding $\theta$, we scale the loss by factor $k$ to reduce the influence of outliers:
\begin{equation}
L(\Delta g_{\text{pred}}, \Delta g_{\text{gt}}) = \begin{cases}
0.5 \cdot \frac{(\Delta g_{\text{pred}} - \Delta g_{\text{gt}})^2}{\beta} & \text{if } |\Delta g_{\text{pred}} - \Delta g_{\text{gt}}| < \beta,\\
|\Delta g_{\text{pred}} - \Delta g_{\text{gt}}| - 0.5 \cdot \beta & \text{if } \beta \leq |\Delta g_{\text{pred}} - \Delta g_{\text{gt}}| < \theta,\\
k \cdot (|\Delta g_{\text{pred}} - \Delta g_{\text{gt}}| - 0.5 \cdot \beta) & \text{if } |\Delta g_{\text{pred}} - \Delta g_{\text{gt}}| \geq \theta
\end{cases}
\label{eq:loss}
\end{equation} 
We use $\beta = \theta = 0.1$ and $k = 0.1$, downweighting outliers by a factor of $10$.

\textbf{Testing.} For testing, we randomly select a fixed number of binocular inputs per subject to serve as the \textit{anchor set} (also referred to as the calibration set)---a small collection of reference images with known gaze targets that enables personalization without retraining. Given a test input from a subject, we compute the gaze displacement $\Delta g_c$ between the input and each image in the anchor set. The predicted gaze is then calculated as shown in Equation~\ref{eq:pred}:
\begin{equation}
    \hat{g} = \frac{1}{C}\sum_{c=1}^{\mathcal{C}} \, (\Delta g_c+g_c)
    \label{eq:pred}
\end{equation}
where $\mathcal{C}$ is the number of calibration samples, $g_c$ is the known gaze target for the $c$-th calibration sample, $\Delta g_c$ is the predicted gaze displacement between the test input and the $c$-th calibration sample. In Section~\ref{sec:Experiments}, we explore the impact of the number of calibration samples on prediction performance.

\section{Experiments}
\label{sec:Experiments}

\subsection{Experiment Setting}  

\noindent \textbf{Dataset and Demographics} 
From the dataset in~\cite{pet_paper}, we use 338 subjects, with 196 for training 
and 142 for validation. The polarization-sensitive camera is positioned to capture 
the top part of the eye, which presents a challenging scenario as this region is 
more prone to occlusions from eyelids and eyelashes. Figure~\ref{fig:data} demonstrates the polarization data processed into three channels: Intensity, DoLP, and AoLP. For personalization experiments, we evaluate two input modalities: \textit{Polarization data} using Intensity-DoLP-AoLP channels, and \textit{Intensity data} using three replicated Intensity channels to ensure fair model capacity comparison. The image dimensions are $3 \times 256 \times 256$ for both modalities. 

Out of all participants, the majority identified as White (52.5\%), followed by Asian (28.0\%), which includes East Asian, South Asian, and Southeast Asian participants. Black or African American participants comprised 6.8\% of the sample, while Hispanic or Latino participants represented 4.5\%. The remaining 8.2\% included individuals identifying as Middle Eastern, American Indian, Native Hawaiian, or Other ethnicities. The data collection protocol permitted the use of contact lenses but absence or presence of contacts is not recorded or used for the purposes of this study. 

\noindent \textbf{Comparing methods.}
We compare the Siamese personalization approach against a Baseline model, where the Baseline corresponds to one branch of the Siamese network that directly produces absolute gaze predictions without user-specific calibration. Due to the shared (Siamese) weights, this single branch has the same model capacity as the full network. Furthermore, we explore the use of linear calibration method on both Siamese and Baseline models, following~\cite{8920005, aria}. Linear calibration is a simple linear model for user-specific calibration. Given the model output $\mathbf{\hat{g}} = [\hat{\theta}_{\text{left}}, \hat{\varphi}_{\text{left}}, \hat{\theta}_{\text{right}}, \hat{\varphi}_{\text{right}}]$, the calibrated output is:

$$\mathbf{\hat{g}_{final}} = \mathbf{\theta_0} + \mathbf{\mu} \odot \mathbf{\hat{g}}$$

where $\mathbf{\theta_0} = [\theta_{0,\text{left}}, \varphi_{0,\text{left}}, \theta_{0,\text{right}}, \varphi_{0,\text{right}}]$ is the bias vector, $\mathbf{\mu} = [\mu_{\theta_{\text{left}}}, \mu_{\varphi_{\text{left}}}, \mu_{\theta_{\text{right}}}, \mu_{\varphi_{\text{right}}}]$ is the scale vector, and $\odot$ denotes element-wise multiplication.

The linear parameters $(\theta_{0,\text{left}}, \mu_{\theta_{\text{left}}}, \varphi_{0,\text{left}}, \mu_{\varphi_{\text{left}}}, \theta_{0,\text{right}}, \mu_{\theta_{\text{right}}}, \varphi_{0,\text{right}}, \mu_{\varphi_{\text{right}}})$ are estimated using all frames in the subject's calibration sequence by minimizing the L1 loss. For this linear calibration process, in~\cite{aria} the whole calibration sequence is utilized which consists of approximately 1000 eye gaze points in their case. In our experiments, linear calibration was done on a sequence of $\sim$100 frames, with 9 target positions arranged in a ring covering 20 degrees.

We compare the Siamese personalization approach against the linearly calibrate Baseline model with PET. Additionally, we explore combining both methods, where personalized gaze predictions from the Siamese model are refined through linear calibration as a post-processing step.

\noindent \textbf{Evaluation Metrics.} 
To evaluate model performance, we use gaze error in degrees as the performance metric. We report different percentiles of the error distribution: P50 (50th percentile/median), P75 (75th percentile), and P95 (95th percentile). While P50 is a commonly reported metric in previous studies, for AR/VR applications with continuous usage, P95 captures the large low-frequency misses that are the main driver of user experience; therefore, we also provide P95 metrics in our analysis.

\noindent \textbf{Feature Extractor.} 
For the experiments, we use the model structure described in~\cite{aria} as the feature extractor. The feature extractor has two independent branches: one processes the left eye image and the other processes the right eye image. Each branch independently extracts features, which are then concatenated for further processing. In the original model of~\cite{aria}, the resulting feature vector is used to directly predict absolute polar coordinates. In contrast, for the Siamese architecture, we use the same pipeline to extract features from another left-right eye image pair from the same subject. The two feature vectors are concatenated and fed to a regressor that outputs the relative gaze difference.

\subsection{Results}

Table~\ref{tab:siamese:baseline} presents the performance of Siamese and Baseline models with and without linear calibration on Polarization (Intensity-DoLP-AoLP) and Intensity-only inputs. The best performance is achieved by combining Polarization inputs with Siamese personalization and subsequent linear calibration. We compare the Siamese personalization using only 9 user-specific calibration images against the Baseline model calibrated on the full calibration sequence of approximately 100 images. Remarkably, Siamese personalization either outperforms or performs at the same level as the linearly calibrated Baseline, demonstrating that effective personalization can be achieved with significantly fewer calibration samples. Specifically, Siamese personalization reduces the gaze error by 5.4\% for P95 compared to the linearly calibrated Baseline, while performing comparably for P50 and P75 with 10 times fewer images.

When larger calibration datasets are available, the Siamese approach enables further refinement through linear calibration. Table~\ref{tab:siamese:baseline} shows that applying linear calibration to Siamese predictions yields additional performance gains. Applying linear calibration to the Siamese model instead of the Baseline model reduces gaze error by 13\% for P50, 11\% for P75, and 8.6\% for P95. Thus, calibrated Siamese predictions consistently outperform calibrated Baseline predictions, confirming the advantage of the Siamese architecture.

Notably, across all error percentiles, Polarization inputs consistently yield lower gaze errors than Intensity inputs, confirming that polarization information provides richer features for gaze estimation compared to traditional intensity-based representations. Compared to the conventional linearly calibrated Baseline model with Intensity data, the proposed Siamese personalization model with Polarization data achieves 27\% improvement at P50, 25\% at P75, and 19\% at P95. When using the Siamese model with linear calibration, Polarization inputs reduce gaze error by 12\% for P50, 12\% for P75, and 5.5\% for P95 compared to Intensity-only inputs. Even without additional calibration, the Siamese model with Polarization inputs achieves error reductions of 12\% for P50, 12\% for P75, and 6.5\% for P95 relative to Intensity-only inputs.

\begin{table}[t]
\centering
\caption{Test gaze error percentiles in degrees (P50, P75, P95) comparing Siamese and Baseline models on Polarization (Intensity-DoLP-AoLP) and Intensity data. Siamese personalization outperforms linearly calibrated Baseline, with the combination of Siamese personalization and linear calibration achieving the best overall performance. The proposed Siamese model with Polarization data achieves lower gaze errors than the conventional linearly calibrated Baseline model with Intensity data.}
\label{tab:siamese:baseline}
\begin{tabular}{@{}lccccc@{}}
\toprule
 &  & \textbf{Linear} & \multicolumn{3}{c}{\textbf{Gaze Error ($^\circ$) $\downarrow$}} \\
\cmidrule(lr){4-6}
\textbf{Data} & \textbf{Model} & \textbf{Calibration} & \textbf{P50} & \textbf{P75} & \textbf{P95} \\
\midrule
\multirow{4}{*}{Polarization}
  & \multirow{2}{*}{Siamese}  & \checkmark & \textbf{0.91} & \textbf{1.51} & \textbf{2.88} \\
  &                           & \texttimes & 1.08 & 1.65 & 2.98 \\
  \cmidrule{2-6}
  & \multirow{2}{*}{Baseline} & \checkmark & 1.05 & 1.69 & 3.15 \\
  &                           & \texttimes & 2.36 & 3.18 & 4.78 \\
\midrule
\multirow{4}{*}{Intensity}
  & \multirow{2}{*}{Siamese}  & \checkmark & 1.03 & 1.72 & 3.05 \\
  &                           & \texttimes & 1.23 & 1.87 & 3.19 \\
  \cmidrule{2-6}
  & \multirow{2}{*}{Baseline} & \checkmark & 1.24 & 2.02 & 3.56 \\
  &                           & \texttimes & 2.77 & 3.67 & 5.32 \\
\bottomrule
\end{tabular}
\end{table}

As described in Section~\ref{sec:siamese}, anchor images are used during inference with Siamese model. Table~\ref{tab:calibration} presents a study varying the number of anchors (3, 5, 7, and 9 calibration images). Results show that increasing the number of anchors consistently reduces gaze errors for both Polarization and Intensity data. Notably, while 9 anchors outperform linear calibration with approximately 100 images, even 3 anchors provide competitive performance, significantly reducing calibration requirements.

\begin{table}[t]
\centering
\caption{Effect of calibration/anchor image count on P50, P75, and P95 percentile errors in the Siamese pipeline. Performance improves with additional calibration images, with 9 images achieving competitive results compared to linear calibration that utilizes the entire calibration sequence. With Polarization data, both the Siamese personalization and linearly calibrated Baseline models outperform the conventional Intensity-based linearly calibrated Baseline.}
\label{tab:calibration}
\begin{tabular}{@{}llcccc@{}}
\toprule
&  & \textbf{Calibration} & \multicolumn{3}{c}{\textbf{Gaze Error ($^\circ$) $\downarrow$}} \\
\cmidrule(lr){4-6}
\textbf{Data} & \textbf{Model} & \textbf{Images} & \textbf{P50} & \textbf{P75} & \textbf{P95} \\
\midrule
\multirow{5}{*}{Polarization}
  & \multirow{4}{*}{Siamese}  & 3  & 1.16 & 1.76 & 3.10 \\
  &                           & 5  & 1.12 & 1.71 & 3.04 \\
  &                           & 7  & 1.11 & 1.68 & 3.01 \\
  &                           & 9  & 1.08 & \textbf{1.65} & \textbf{2.98} \\
  \cmidrule{2-6}
  & Baseline                 & $\sim$ 100 & \textbf{1.05} & 1.69 & 3.15 \\
\midrule
\multirow{5}{*}{Intensity}
  & \multirow{4}{*}{Siamese}  & 3  & 1.34 & 2.02 & 3.34 \\
  &                           & 5  & 1.27 & 1.94 & 3.27 \\
  &                           & 7  & 1.26 & 1.91 & 3.24 \\
  &                           & 9  & \textbf{1.23} & \textbf{1.87} & \textbf{3.19} \\
  \cmidrule{2-6}
  & Baseline                 & $\sim$ 100 & 1.24 & 2.02 & 3.56 \\
\bottomrule
\end{tabular}
\end{table}

\begin{figure}[htbp]
    \centering
    \begin{subfigure}[b]{0.45\textwidth}
        \centering
        \includegraphics[width=\textwidth]{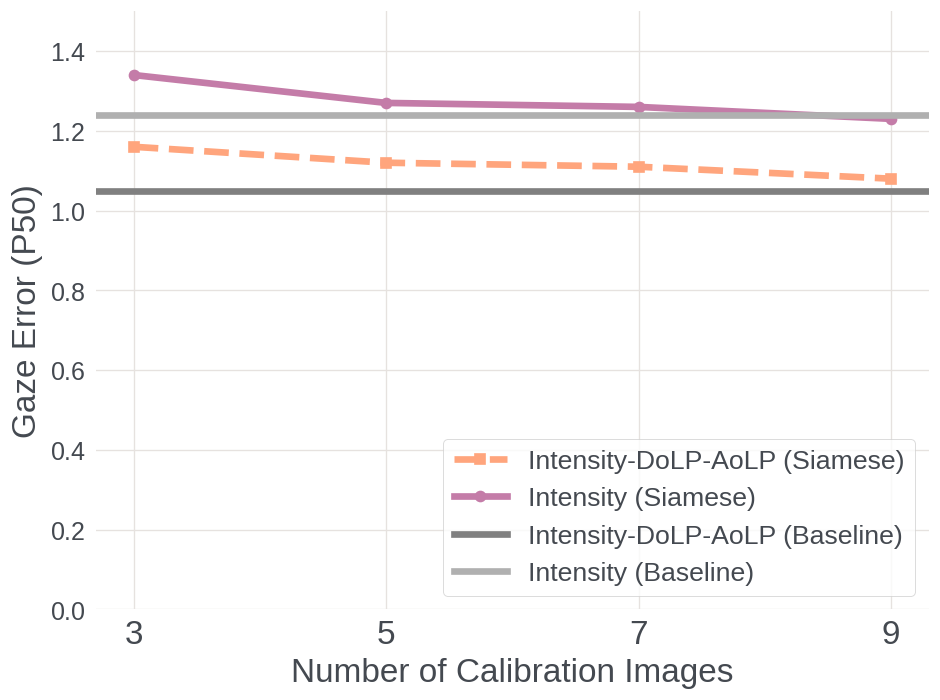}
        \caption{P50 (50th percentile/median) of the error distribution}
        \label{fig:number_anchors_p50}
    \end{subfigure}
    \hfill
    \begin{subfigure}[b]{0.45\textwidth}
        \centering
        \includegraphics[width=\textwidth]{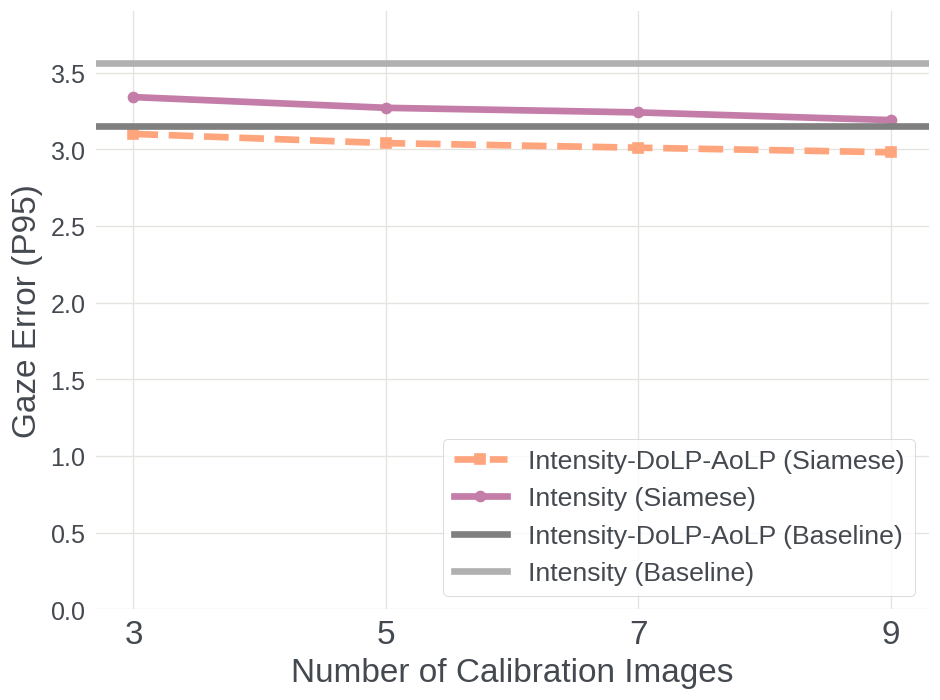}
        \caption{P95 (95th percentile) of the error distribution}
        \label{fig:number_anchors_p95}
    \end{subfigure}
    \caption{P50 (left) and P95 (right) gaze error as a function of the number of anchor images used during Siamese model inference. With 9 anchors, performance matches the Baseline model calibrated on $\sim$100 images.}
    \label{fig:number_anchors}
\end{figure}

Figure~\ref{fig:number_anchors} shows the evolution of median (P50) and P95 test gaze error for the Siamese model with varying numbers of calibration images (3, 5, 7, 9), compared against the Baseline model with linear calibration on approximately 100 images (horizontal lines). For both models, Polarization input consistently outperforms Intensity. As the number of anchor images increases, test gaze error decreases for both input modalities. Notably, Siamese personalization with only 9 images achieves median gaze error comparable to the Baseline model calibrated on the full sequence of approximately 100 images, as shown in Figure~\ref{fig:number_anchors_p50}. Moreover, Figure~\ref{fig:number_anchors_p95} shows that even with just 3 calibration images, Siamese personalization outperforms the linearly calibrated Baseline at P95, indicating superior performance on challenging cases with reduced tail errors. Table~\ref{tab:calibration} further supports that this advantage of Siamese personalization with fewer calibration images persists across all error percentiles (P50, P75, and P95) compared to linear calibration with the Baseline model using $\sim$100 images.

\subsubsection{Ablation Study}
\subsubsection*{\textbf{Training Data Sampling Strategy}}
\begin{figure*}[htbp]
    \centering
    \begin{subfigure}[b]{0.44\textwidth}
        \centering
        \includegraphics[width=\textwidth]{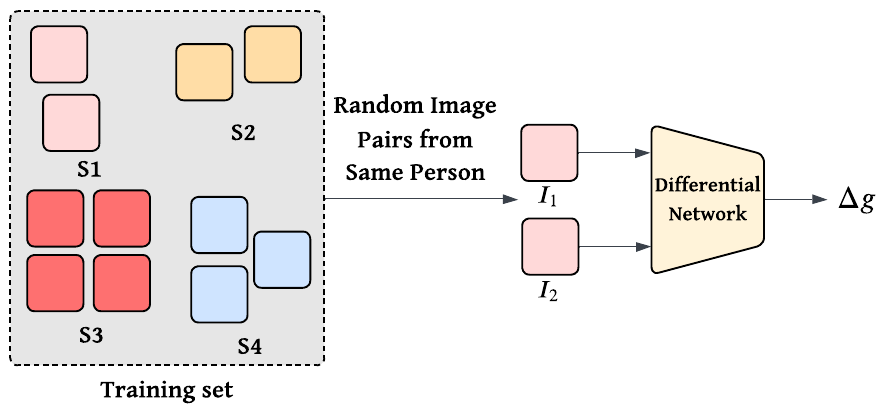}
        \caption{Random Sampling}
        \label{fig:random_sampling}
    \end{subfigure}
    \hfill
    \begin{subfigure}[b]{0.54\textwidth}
        \centering
        \includegraphics[width=\textwidth]{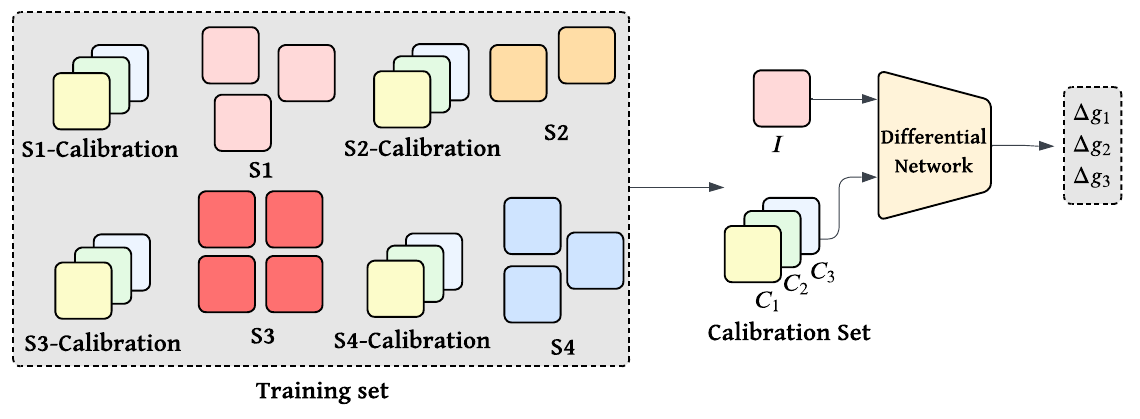}
        \caption{Calibration Sampling}
        \label{fig:calibration_sampling}
    \end{subfigure}
    \caption{Input sampling strategies for training. Random sampling generates random input pairs for each subject, whereas calibration sampling pairs each input with calibration images from a fixed anchor set selected for that subject.}

    \label{fig:training_sampling}
\end{figure*}

Siamese training requires paired inputs as described in Section~\ref{sec:siamese}. In addition to the random sampling strategy, we explore an alternative training data sampling approach which we refer to as \textit{calibration sampling}. Figure~\ref{fig:training_sampling} shows the two sampling strategies: random sampling and calibration sampling. In this figure, S denotes a subject, I denotes an input sample, and $C$ denotes a sample from the subject's calibration set. Each input consists of a left-right eye image pair. Figure~\ref{fig:random_sampling} show random sampling, where we pair inputs from the same training subject randomly. For calibration sampling shown in Figure~\ref{fig:calibration_sampling}, we randomly select 3 calibration images per subject at the beginning to form a fixed calibration set. Training pairs consist of one input from a random time step and from the fixed calibration set for the same subject.

Table~\ref{tab:sampling} shows that, for Polarization, random sampling yields lower test errors, suggesting better generalization. The fixed calibration images were randomly selected per subject; however, this strategy may encourage memorization rather than generalization.

\begin{table}[t]
\centering
\caption{Performance comparison of random sampling vs. calibration sampling (with 3 Images/Subject) for Siamese model training on Polarization data. Random sampling shows lower gaze errors than pre-selected fixed calibration sets.}
\label{tab:sampling}
\begin{tabular}{@{}llccc@{}}
\toprule
 &  & \multicolumn{3}{c}{\textbf{Gaze Error}} \\
\cmidrule(lr){3-5}
\textbf{Data} & \textbf{Sampling Strategy} & \textbf{P50} & \textbf{P75} & \textbf{P95} \\
\midrule
\multirow{2}{*}{Polarization}
  & Random Sampling & \textbf{1.08} & \textbf{1.65} & \textbf{2.98} \\
  \cmidrule{2-5}
  & Calibration Sampling & 1.34 & 2.05 & 3.56 \\
\bottomrule
\end{tabular}
\end{table}

\subsubsection*{\textbf{Intensity vs. Single-channel Intensity}} 
While Intensity data typically has a single channel, we explore triplicate Intensity (three identical Intensity channels) to ensure fair comparison with 3-channel Polarization data (Intensity-DoLP-AoLP). Table~\ref{tab:triplicate_intensity} presents three configurations with Baseline model: single-channel Intensity, triplicate Intensity, and Polarization. Triplicate Intensity and single-channel Intensity achieve similar performance when linear calibration is applied; however, without linear calibration, triplicating the Intensity channel outperforms single-channel Intensity. Therefore, we use triplicate Intensity as the Intensity-based Baseline in all comparisons and refer to it as the Intensity experiment. While triplicating the Intensity channel improves performance over single-channel input, likely due to increased model capacity, Polarization consistently outperforms Intensity when linear calibration is applied, reducing error by 15\% for P50, 16\% for P75, and 12\% for P95. So, even though we observe performance improvement due to the increase in the number of channels, we obtain the best results with Polarization, suggesting that the performance gains are not solely due to increased channel capacity but also from the polarization information itself.

\begin{table}[t]
\centering
\caption{P50, P75, and P95 percentile errors for different input modalities. Triplicating the Intensity channels reduces error compared to single-channel Intensity input, but Polarization achieves the best performance, demonstrating that the improvement is due to polarization features rather than increased channel count.}
\label{tab:triplicate_intensity}
\begin{tabular}{@{}lcccc@{}}
\toprule
 & \textbf{Linear} & \multicolumn{3}{c}{\textbf{Gaze Error ($^\circ$) $\downarrow$}} \\
\cmidrule(lr){3-5}
\textbf{Data} & \textbf{Calibration} & \textbf{P50} & \textbf{P75} & \textbf{P95} \\
\midrule
\multirow{2}{*}{Polarization}
  & \checkmark & \textbf{1.05} & \textbf{1.69} & \textbf{3.15} \\
  & \texttimes & 2.36 & 3.18 & 4.78 \\
\midrule
\multirow{2}{*}{Intensity}
  & \checkmark & 1.24 & 2.02 & 3.56 \\
  & \texttimes & 2.77 & 3.67 & 5.32 \\
\midrule
\multirow{2}{*}{Single-channel Intensity}
  & \checkmark & 1.26 & 2.04 & 3.57 \\
  & \texttimes & 2.83 & 3.74 & 5.38 \\
\bottomrule
\end{tabular}
\end{table}

\subsection{Discussion and Practical Considerations}
\label{sec:discussion}

\paragraph{Headset Slippage.}
In real-world deployment, headset slippage can introduce systematic gaze offsets that degrade tracking accuracy over time. Siamese personalization offers a natural mechanism for re-calibration: when slippage is detected (e.g., via IMU signals or user-initiated recalibration), the system can rapidly update the anchor set with minimal user effort, leveraging the approach's efficiency with few calibration samples.

\paragraph{Computational Latency.}
The Siamese architecture requires one forward pass per anchor image during inference. For our 9-anchor configuration, this results in approximately 9× the computational cost of the Baseline model per gaze prediction. In latency-critical applications, this overhead can be mitigated by caching anchor features after calibration, reducing inference to a single forward pass plus lightweight displacement aggregation. Alternatively, reducing the anchor count to 3--5 images maintains competitive accuracy while lowering computational burden, as shown in Table~\ref{tab:calibration}.

\section{Conclusion and Future Work}
\label{sec:Conclusion}
In this work, we demonstrated that Siamese personalization can be effectively applied to Polarization-enabled Eye Tracking (PET) using only binocular eye images as input, achieving high accuracy while maintaining low calibration burden suitable for real-world deployment. Our results reveal that Siamese personalization with Polarization data (Intensity-DoLP-AoLP) requires 10× fewer calibration images than traditional linear calibration method applied to the Baseline model while achieving comparable or superior performance. Furthermore, we showed that Siamese personalization and linear calibration are complementary approaches: combining both methods yields the best overall performance, with error reductions of 13\% at P50, 11\% at P75, and 8.6\% at P95 compared to applying linear calibration to the Baseline model. This finding enables practical deployment where systems achieve accurate gaze estimation with minimal calibration through Siamese personalization, with the flexibility to further refine predictions as more calibration data becomes available. Future work could explore intelligent calibration data selection strategies to identify the most informative images for each user's calibration set, balancing challenging cases where the model exhibits high uncertainty with representative examples that serve as reliable anchors for robust personalization. Additionally, comparisons with meta-learning approaches such as MAML would provide valuable theoretical context for isolating the specific contributions of the Siamese architecture; however, such methods have not yet been validated on polarization-enabled eye tracking data and represent a promising direction for future investigation.

\bibliographystyle{assets/plainnat}
\bibliography{paper}

\end{document}